\documentclass[letterpaper,10pt,conference]{ieeeconf}

\IEEEoverridecommandlockouts                          

\title{\LARGE \bf
DockAnywhere: Data-Efficient Visuomotor Policy Learning for Mobile Manipulation via Novel Demonstration Generation
}
\author{Ziyu Shan$^{1}$, Yuheng Zhou$^{1}$, Gaoyuan Wu$^{1}$, Ziheng Ji$^{1}$, Zhenyu Wu$^{2}$ and Ziwei Wang$^{1 \dagger}$
\thanks{$^{1}$ Nanyang Technological University, Singapore}
\thanks{$^{2}$ Beijing University of Posts and Telecommunications, Beijing, China}
\thanks{$^{\dagger}$Corresponding author: \textit{ziwei.wang@ntu.edu.sg}}
}
\usepackage{graphicx} 
\usepackage{amsmath}
\usepackage{multirow}
\usepackage{multicol}
\usepackage{booktabs}
\usepackage{bbding}
\usepackage{makecell}
\usepackage{cite}
\usepackage{url}
\usepackage{tabu, booktabs}
\usepackage[table]{xcolor}
\newcommand{\cmrwordtitle}[1]{{\fontsize{8pt}{8pt}\textbf{\texttt{#1}}}}
\newcommand{\cmrword}[1]{{\fontsize{8.7pt}{8pt}{$\mathtt{#1}$}}}
\newcommand{\cmrwordsmall}[1]{{\fontsize{8pt}{8pt}{$\mathtt{#1}$}}}
\newcommand{\ans}[1]{\textcolor[RGB]{0, 0, 0}{#1}}

\usepackage{hyperref}
\begin{document}

\maketitle
\thispagestyle{empty}
\pagestyle{empty}

\begin{abstract}
Mobile manipulation is a fundamental capability that enables robots to interact in expansive environments such as homes and factories. Most existing approaches follow a two-stage paradigm, where the robot first navigates to a docking point and then performs fixed-base manipulation using powerful visuomotor policies. However, real-world mobile manipulation often suffers from the view generalization problem due to shifts of docking points. To address this issue, we propose a novel low-cost demonstration generation framework named \small{\cmrwordtitle{DockAnywhere}}, which improves viewpoint generalization under docking variability by lifting a single demonstration to diverse feasible docking configurations. Specifically, \cmrwordtitle{DockAnywhere} lifts a trajectory to any feasible docking points by decoupling docking-dependent base motions from contact-rich manipulation skills that remain invariant across viewpoints. Feasible docking proposals are sampled under feasibility constraints, and corresponding trajectories are generated via structure-preserving augmentation. Visual observations are synthesized in 3D space by representing the robot and objects as point clouds and applying point-level spatial editing to ensure the consistency of observation and action across viewpoints. Extensive experiments on ManiSkill and real-world platforms demonstrate that \cmrwordtitle{DockAnywhere} substantially improves policy success rates and easily generalizes to novel viewpoints from unseen docking points during training, significantly enhancing the generalization capability of mobile manipulation policy in real-world deployment. The code can be found at \url{https://github.com/zyshan0929/DockAnywhere}.
\end{abstract}

\section{INTRODUCTION}
Mobile manipulation emerges as a key milestone toward integrating robotics into everyday life, which requires robots to perform complex manipulation tasks across expansive environments~\cite{pankert2020perceptive,haviland2022holistic,li2025momagen,wu2025moto,jang2023motion}. 
Inspired by the tremendous advances of fixed-base manipulation with impressive capabilities such as visuomotor policies~\cite{yang2025mobi,chai2025n2m} and vision-and-language models~\cite{wu2025momanipvla,gubernatorov2025anywherevla,lin2025echovla}, existing approaches try to extend these successes to mobile manipulation by following a two-stage paradigm: first navigating to a docking point and then performing fixed-base manipulation with the assistance of powerful manipulation policies~\cite{yin2024sg, fang2023anygrasp}.

\begin{figure}[t]
    \centering
    \includegraphics[width=1.0\columnwidth]{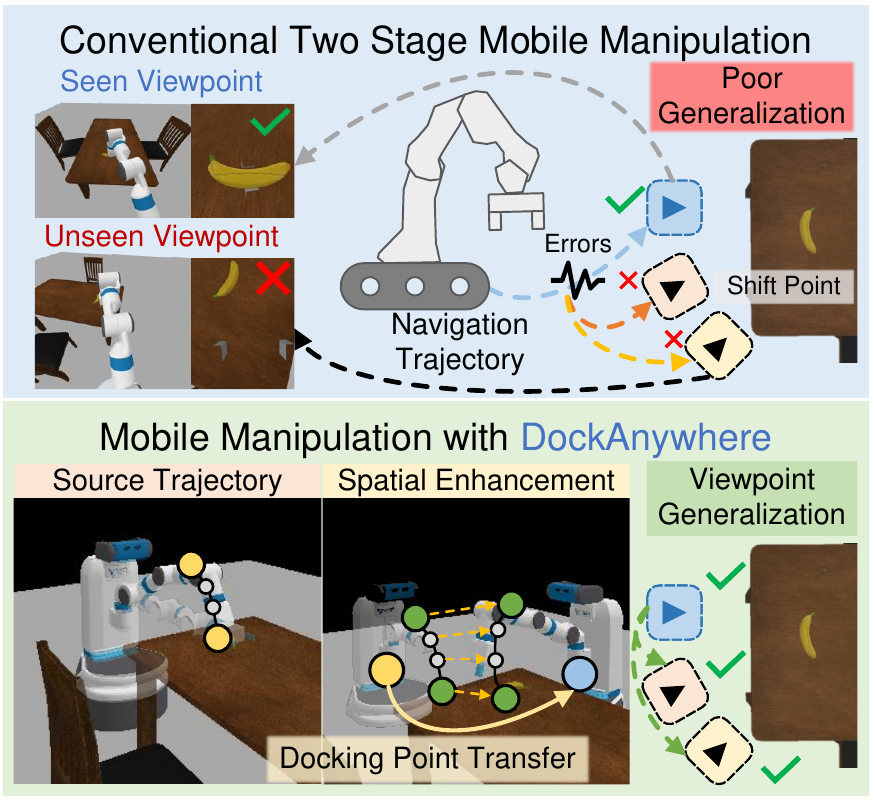} 
    \caption{Docking point shift in conventional two-stage mobile manipulation causes severe viewpoint variation, presents challenges for fixed-base manipulation. Our \cmrwordsmall{DockAnywhere} utilizes trajectory-based viewpoint augmentation to generate diverse viewpoints from a single trajectory, ensuring that the manipulation policy can generalize to novel viewpoints caused by navigation errors.}
    \label{fig:teaser}
    \vspace{-0.6cm}
\end{figure}

However, navigation in real-world environments is subject to errors including sensing noise and localization bias, which often prevent the robot from reaching the intended docking pose and result in inevitable docking point shifts.
Meanwhile, the shift of the docking point can lead to significant changes of the robot's egocentric visual observations, resulting in a severe \textbf{view generalization problem} for the subsequent fixed-base manipulation policy.
As illustrated in Fig.~\ref{fig:teaser}, this fundamental bottleneck of the view generalization problem caused by navigation errors limits the performance and robustness of mobile agents, particularly in unpredictable real-world environments where the agents need to navigate to novel spatial configurations while maintaining manipulation precision \cite{yang2023movie,yuan2024learning}.

Several recent mobile manipulation approaches~\cite{zhu2025emma,yuan2025hermes,li2025momagen} try to indirectly address the view generalization problem by formulating novel trajectory generation as an optimization problem to expand training data and searching for base poses and viewpoints that are better conditioned for perception.
These approaches aim to improve training coverage by restricting feasible viewpoints or base pose using heuristic constraints such as visibility or reachability. 
While effective to some extent, existing approaches rely on limited offline data and handcrafted constraints, which limit their ability to generalize when navigation errors lead to previously unseen docking configurations at test time.

To address the issue, we argue that the choice of visual representation is a central factor in addressing docking-induced viewpoint variation.
Instead of relying on egocentric observations, we adopt a third-person viewpoint representation as the fixed-base manipulation policy input, which provides a consistent scene representation across different docking points. 
\ans{Such third-person observations are readily available in many real-world scenarios, where the robot often operates in functionally constrained areas such as kitchens. In these settings, third-person viewpoints enables a practical sensing modality with global and less-occluded overview of the workspace.}
In the context of mobile manipulation, this representation substantially reduces appearance variation caused by navigation errors, effectively transforming the viewpoint generalization challenge into a geometric docking point generalization problem, where the policy primarily needs to reason over changes in the robot base pose rather than over drastically altered visual observations.

Building on this insight, we propose a low-cost and fully synthetic demonstration generation framework named \cmrword{DockAnywhere}, which explicitly targets viewpoint generalization under docking point variability in mobile manipulation.
Inspired by recent advances in trajectory generation for fixed-base manipulation~\cite{mandlekar2023mimicgen,garrett2024skillmimicgen,xue2025demogen}, \cmrword{DockAnywhere} integrates task and motion planning (TAMP)~\cite{mandlekar2023human} as a structural constraint for trajectory augmentation.
Specifically, a source manipulation trajectory is explicitly parsed into two semantically distinct components: low-precision \textit{motion} segments, which primarily encode fixed-base approach movements of robot arms, and contact-rich \textit{skill} segments, which capture fine-grained object interactions.
This decomposition enables \cmrword{DockAnywhere} to lift a single demonstration to any feasible docking points.
For novel docking configurations, \cmrword{DockAnywhere} relocates the robot base and regenerates corresponding \textit{motion} segments through TAMP, while preserving entire \textit{skill} segments to maintain contact consistency and interaction fidelity. In this way, docking point variation is absorbed by motion-level adaptation, while manipulation skills remain invariant.
Crucially, \cmrword{DockAnywhere} further ensures cross-viewpoint consistency between actions and observations by synthesizing visual inputs through 3D point-level editing.
The robot and objects are represented as point clouds and transformed using the same spatial mappings applied to trajectory augmentation, yielding physically and visually coherent demonstrations that provide high-quality supervision for learning spatially robust manipulation policies.

Extensive experiments on ManiSkill \cite{taomaniskill3} demonstrate that \cmrword{DockAnywhere} substantially improves policy performance of the original DP3 \cite{Ze2024DP3} and fixed-base manipulation augmentation framework like DemoGen \cite{xue2025demogen} in terms of average success rates, even on the unseen docking points at test time.
Additionally, we conduct multiple real-world tasks, showing that our \cmrword{DockAnywhere} can be deployed easily on a mobile manipulator platform with only a third-person camera for observation. With the negligible computation cost of $\sim$0.1 seconds for all augmentation based on one source trajectory, \cmrword{DockAnywhere} substantially enhances viewpoint generalization capability, generalizing to unseen docking points and achieving an average success rate of 43.3\% using a limited set of source demonstrations.  Finally, we demonstrate that \cmrword{DockAnywhere} enables the deployed mobile policy to operate even with a out-of-distribution distance from the workspace, showing the potential to address the view generalization problem of mobile manipulation by sufficiently training the policy in a low-cost synthesis approach.

\section{RELATED WORK}
\textbf{Learning for Mobile Manipulation.} Prior work has explored various learning-based approaches for mobile manipulation. Some methods rely on end-to-end learning but inevitably require large-scale training data. 
In contrast, the modular mobile manipulation framework utilizes a two-stage paradigm: off-the-shelf navigation followed by fixed-base manipulation. 
ARPlace \cite{stulp2012learning} predicts future robot poses for running skills but concentrate on reachability rather than viewpoint restrictions for these skills. 
MoManipVLA \cite{wu2025momanipvla} achieves data-efficient generalized mobile manipulation via optimization of the base waypoints to ensure that VLA prediction trajectories are feasible. 
However, while real-world navigation frequently suffers from suboptimal or even perturbed docking locations, resulting in view generalization problem and limiting the performance of subsequent fixed-base manipulation.

Pioneering works have developed new data scaling frameworks that transform humans demonstrations into mobile manipulation data \cite{zhu2025emma,yuan2025hermes,li2025momagen} to improve the generalization capability of mobile agents, including the performance across multiple docking points. 
EMMA \cite{zhu2025emma} develops a new data collection framework that that leverages human full-body motion data and static robot data to fill in the gap of data scarcity across different navigation destinations.
HERMES \cite{yuan2025hermes} effectively leverages diverse human motion data sources and integrate them with robust sim2real methodologies by a hybrid control scheme alongside a generalized distillation framework.
MoMaGen \cite{li2025momagen} proposes a general data generation method for multi-step bimanual mobile manipulation. The proposed framework formulates the generation as a constrained optimization problem subject to hard constraints while balancing soft constraints.
However, these methods require expensive hardware facilities to collect human motion data and are inevitably labor-intensive, still failing to address the fundamental bottleneck of data scarcity across multiple docking points.
\begin{figure*}
    \centering
    \includegraphics[width=1\textwidth]{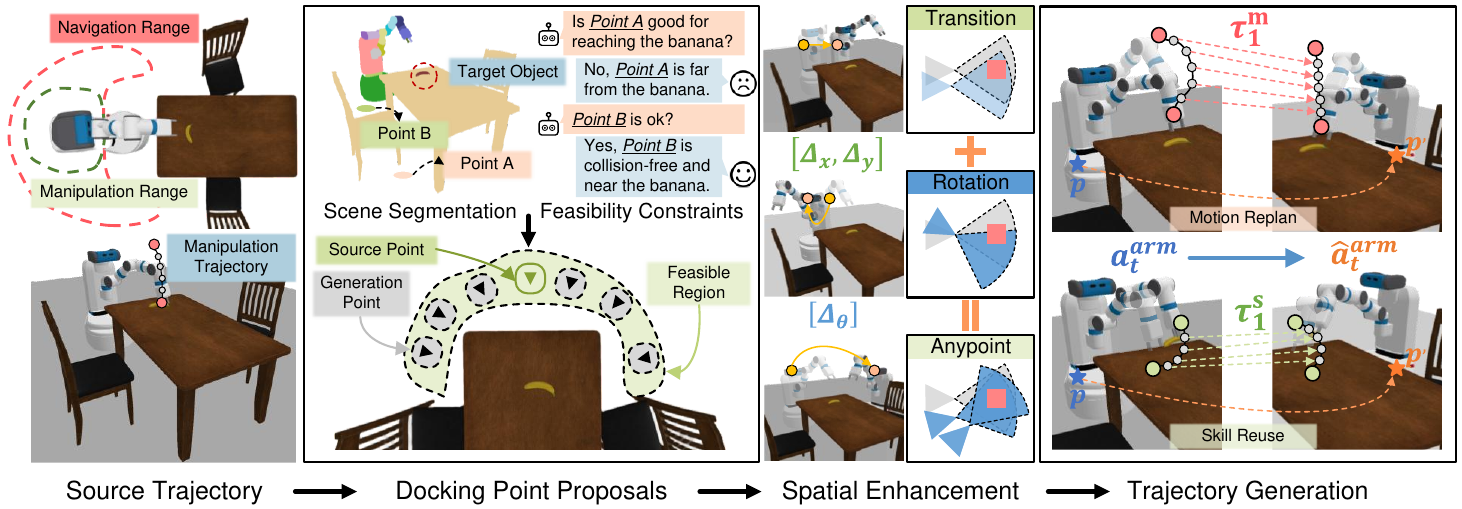} 
    \caption{Pipeline of the \cmrwordsmall{DockAnywhere} framework. Given one source demonstration, \cmrwordsmall{DockAnywhere} captures the point cloud and parses the trajectory based on the segmentation results. TAMP-based spatial transformation is applied to generate actions for relocated robot. Then the visual observations are synthesized by perceiving the robot's end-effector and objects as 3D point clouds and relocating the position of the robot via point-level editing.}
    \label{fig:pipeline}
    \vspace{-0.55cm}
\end{figure*}

\textbf{Data-Efficient Imitation Learning.} Data-efficient imitation learning~\cite{chen2023genaug,ameperosa2025rocoda,yuan2025roboengine,valassakis2022demonstrate,wen2022you,johns2021coarse} attempts to learn useful manipulation policies from limited amount of demonstrations.
VISTA \cite{tian2024view} proposes synthesizing novel camera views from single-view demonstrations, allowing a policy to train on visually diverse observations while preserving the underlying action semantics, while RoVi-Aug \cite{chen2024rovi} extends this idea with joint robot and viewpoint augmentation, generating demonstrations that vary both camera pose and robot embodiment so that policies can generalize across different robot morphologies.
RoCoDA~\cite{ameperosa2025rocoda} introduces counterfactual data augmentation for robot learning from demonstrations by generating causally consistent variations of observed trajectories, improving generalization from limited data.
Another common strategy is introducing imitation learning to replace specific segments in the Task and Motion Planning (TAMP) pipeline. 
DemoGen \cite{xue2025demogen} further eliminates the need for on-robot rollouts by synthesizing visual observations using a pre-trained diffusion model, but it is limited to 2D image editing and cannot handle 3D spatial transformations.
CP-Gen \cite{lin2025constraint} achieves geometry-aware data generation by formulating keypoint-trajectory constraints where keypoints on the robot or grasped object must track a reference trajectory defined relative to a task-relevant object.
However, all these methods primarily focus on static manipulation and can not address the additional complexity of base motion for mobile manipulation tasks.

\section{METHOD}
Aimed at improving the spatial generalization capability of mobile manipulation policy across multiple docking point candidates, \cmrword{DockAnywhere} focuses on generating spatially augmented observation-action sequences from a small set of source demonstrations. 
Specifically, \cmrword{DockAnywhere} first parses the source trajectory and split it into relatively low-precision \textit{motion} segments and \textit{skill} segments, followed by TAMP-based spatial transformation to generate novel trajectories for the relocated docking point.
Afterwards, the visual observations are synthesized by perceiving the robot and objects as 3D point clouds in a point-level editing strategy. 

\subsection{Problem Formulation}
Popular two-stage mobile manipulation methods \cite{wu2025momanipvla,gubernatorov2025anywherevla,yin2024sg} first navigate the robot to a docking point $p$ and then learn the fixed-base manipulation policy $\pi_\text{ego}: \mathcal O_\text{ego} \mapsto \mathcal A_\text{ego}$ to map the \textit{egocentric} visual observations $o_\text{ego} \in \mathcal O_\text{ego}$ to the predicted actions $a_\text{ego} \in \mathcal A_\text{ego}$.
However, these two-stage methods often suffer from the view generalization problem, where small shifts of docking point $p'$ can  result in misalign of the robot's egocentric observation $o'_\text{ego}$ and thus lead to task failure.   
To solve this, we use a \emph{third-person} view $o \in O$ as the robot's observation to avoid the perturbation caused by $o'_\text{ego}$, providing a consistent view across different docking poses. 
In this way, we explicitly convert the non-trivial view generalization problem (\textit{i.e.,} $o_\text{ego} \rightarrow o'_\text{ego}$) into a geometric docking point generalization problem (\textit{i.e.,} $p \rightarrow p'$) by introducing a fixed viewpoint.

Under this third-person view, given the dataset \(\mathcal{D}_{p}=\{(o_t,a_t)|p\}_{t=1}^{T}\) with observation-action pair $(o_t, a_t)$ collected at docking point $p$ and the fixed-base manipulation policy \(\pi\), our goal is to learn a generalizable policy \(\pi_{p'}\) that can predict feasible actions $a'_t$ at the unseen docking point \(p'\):
\begin{equation}
a'_t \sim \pi_{p'}\!\left(\,\cdot \;\middle|\; o'_t,\, a'_{t-1}\right).
\end{equation}
where $\cdot$ denotes the action variable sampled from the conditional policy. Ideally, \(\pi_{p'}\) is obtained by learning from actual demonstrations \(\mathcal{D}_{p'}\) at \(p'\) via behavior cloning. Unfortunately, \(\mathcal{D}_{p'}\) is initially unavailable since \(p'\) is unseen in our scenario. To fill this gap, we aim to generate \(\mathcal{\hat D}_{p'}\) from \(\mathcal{D}_{p}\) via our \cmrword{DockAnywhere} framework by \textit{data augmentation}, enabling the learning of \(\pi'\) through behavior cloning:
\begin{equation}
\min_{\pi_{p
'}}\; \mathcal{L}_{\mathrm{BC}}(\pi_{p'};\mathcal{\hat D}_{p'})
= -\frac{1}{\lvert \mathcal{\hat D}_{p'}\rvert}\sum_{(o'_t,a'_t)\in \mathcal{\hat D}_{p'}}
\log \pi_{p
'}\!\left(a'_t \mid o'_t,\, a'_{t-1}\right).
\end{equation}

Formulating the data augmentation process, our \cmrword{DockAnywhere} generates $\mathcal{\hat D}_{p'}$ from the source demonstration $\mathcal D_{p}$:
\begin{equation}
\mathcal {\hat D}_{p'} = (\hat s_0, \hat s_1,..., \hat s_{L-1}|p')    
\end{equation}
where $\hat s_t = (\hat o_t, \hat a_t)$ denotes the generated observation-action pair at time step $t$.
For the action format, the action $a_t=(a_{t}^{\text{pose}}, a_{t}^{\text{cmd}})$ consists of the robot arm and end-effector (EE) command, where $a_{t}^{\text{pose}}$ is the target EE pose in the world frame, and $a_{t}^{\text{cmd}}$ is the binary signal controlling the gripper's close action. 
The observation $o_t=(o_t^{\text{pc}},o_t^{\text{state}})$ includes the point cloud data and the proprioceptive of the robot, where $o_t^{\text{state}}$ denote the 1D state of the end-effector.
{Note that this design can be easily extended to scenarios where only onboard sensors are available by simple re-projection using calibrated extrinsics. This flexibility is due to the direct augmentation in 3D point-cloud space, providing a natural path toward onboard-only and egocentric policy learning for mobile manipulation.}

\subsection{Demonstration Preprocessing}
\label{sec:preprocess}
Given the source demonstration $\mathcal D_{p}$, \cmrword{DockAnywhere} first preprocesses the trajectory by capturing the point cloud and parsing the trajectory into \textit{motion} and \textit{skill} segments. To generate object-centric \textit{skill} segments, we also need to identify the involved objects using point cloud segmentation.

\noindent\textbf{Point Cloud Acquisition and Segmentation.} Similar to~\cite{Ze2024DP3,xue2025demogen}, we utilize a single-view RGBD camera in the real-world scene for point cloud acquisition. The RGBD image is converted into a raw point cloud in the world frame given the camera parameters~\cite{zhang2024asynchronous}. Since the converted point clouds may contain the redundant points such as the points of the ground and robot's torso, we crop out these points out of a predefined 3D bounding box to focus on the robot's end-effector. Then, we further downsample the cropped point cloud to 1024 or 2048 points by farthest point sampling (FPS) \cite{qi2017pointnet} to reduce the computational overhead, as shown in Fig.~\ref{fig:pipeline}.

As for the segmentation masks of the robot and objects in the point cloud, we apply an off-the-shelf segmentation model (\textit{e.g.}, Grounded SAM~\cite{ren2024grounded}) on the first frame of RGB image to obtain the pixel-aligned segmentation map, which are then projected to the raw point cloud to obtain the 3D segmentation masks. 

\noindent\textbf{Source Trajectory Parsing.} Based on the common observation that the robot must initially approach the target object and then manipulate it through contact, we parse the source trajectory $\mathcal D_{p}$ into a sequence of free-space \textit{motion} and \textit{skill} segments.
Specifically, for each segment with the time span $\tau = [t_{\text{start}}, t_{\text{end}}] \subseteq [0, L)$, its category can be determined by the distance between the robot's end-effector and the geometric center of the target object. \ans{One segment is identified as a \textit{skill} segment if the distance is below a empirically predefined threshold, indicating potential contact with the object.} Otherwise, it is classified as a \textit{motion} segment. Therefore, the source trajectory $\mathcal D_{p}=(d_0,...,d_{L-1}|p)$ can be represented as a sequence of $l$ segments:

\begin{equation}
    \mathcal D_{p} = (d[\tau_1^{\text{m}}], d[\tau_1^{\text{s}}],..., d[\tau_l^{\text{m}}], d[\tau_l^{\text{s}}]|p)
\end{equation}

\begin{figure}
    \centering
    \includegraphics[width=0.48\textwidth]{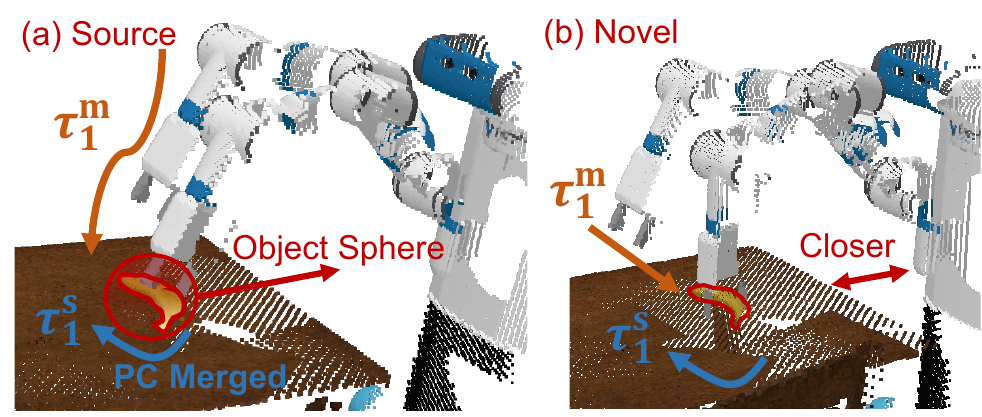} 
    \caption{Illustration of source trajectory parsing and novel trajectory generation for the pick-up banana task. The trajectory is parsed into a sequence of \textit{motion} and \textit{skill} segments based on the distance between the end-effector and the target object. Then the \textit{motion} segments are replanned for the relocated docking point, while the \textit{skill} segment is reused as a whole.}
    \label{fig:parse}
    \vspace{-0.6cm}
\end{figure}

For the instance of pick-up banana task illustrated in Fig.  \ref{fig:parse} (a), the trajectory is parsed into two segments by checking if the position of the end-effector lies in the object sphere: (1) \textit{motion} segment $\tau_1^\text{m}$ where the gripper approaches the banana, (2) \textit{skill} segment $\tau_1^\text{s}$ where the robot grasps and lifts together with the banana.

{Then, to obtain feasible docking point candidates, three constraints have to be satisfied: 1) \textit{visibility}, where the segmented target-object points remain observable; 2) \textit{reachability}, where the target remains inside the robot workspace from the new base pose; and 3) \textit{collision-free}, where each motion segment can be replanned into a valid trajectory without collision. A sampled docking point is accepted only if all motion segments can be successfully replanned into collision-free and feasible trajectories; otherwise, the docking point is discarded.}

\subsection{Action Generation via TAMP-based Transformation}
\noindent\textbf{Action Generation.} As mentioned above, the actions consist of the robot arm and hand commands. For example in Fig.  \ref{fig:parse}, the robot hand commands $a_t^{\text{cmd}}$ are binary signals indicating whether to close the gripper. Obviously, these hand commands are invariant of spatial transformation, thus $a_t^{\text{cmd}}$ can be directly reused for each frame of the trajectory regardless of the docking point relocation:
\begin{equation}
    \hat a_t^{\text{cmd}} = a_t^{\text{cmd}}, \quad  t \in [0, L)
\end{equation}

Conversely, the robot arm commands are equivariant to the spatial transformations according to the altered docking point.
Specifically, regarding the \textit{skill} segments, the spatial relations between the end-effector and the target object must be relatively static to maintain the complex interaction. Therefore, the \textit{skill} segments are reused as a whole for overall spatial transformation. Since the object configuration $s_0$ remains the same as the source, the arm actions in the $k$-th \textit{skill} segments $\tau_k^{\text{s}}$ can be reused as:
\begin{equation}
    \hat a_t^{\text{pose}} =  a_t^{\text{pose}}, \quad  t \in \tau_k^{\text{s}}
\end{equation}

Regarding the \textit{motion} segments, they are relatively low-precision and can be easily replanned for the relocated docking point. Specifically, we utilize a motion planner to replan the arm actions in the $k$-th \textit{motion} segment $\tau_k^{\text{m}}$ from the end pose $a_{t_{\text{end},k-1}}^{\text{pose}}$ of the last skill segment $\tau_{k-1}^{\text{s}}$ to the start pose $a_{t_{\text{start},k}}^{\text{pose}}$ of $\tau_{k}^{\text{s}}$ to connect two adjacent skill segments:
\begin{equation}
    \hat a_t^{\text{pose}} = \mathtt{Replan}(\hat a_{t_{\text{end},k-1}}^{\text{pose}}, \hat a_{t_{\text{start},k}}^{\text{pose}}), t \in \tau_k^{\text{m}}
\end{equation}
Note that the action $a_{t_{\text{end},k-1}}^{\text{pose}}$ is also the target pose of the last frame in $\tau_{k-1}^{\text{s}}$ due to the control mode of target pose, so the skill segments can be seamlessly connected. \ans{When relative location between the robot and object varies, we update the desired end-effector pose for manipulation and preserve the relative position in the object frame, then replan the motion segment to this updated goal.}

\begin{figure*}
    \centering
    \includegraphics[width=0.86\textwidth]{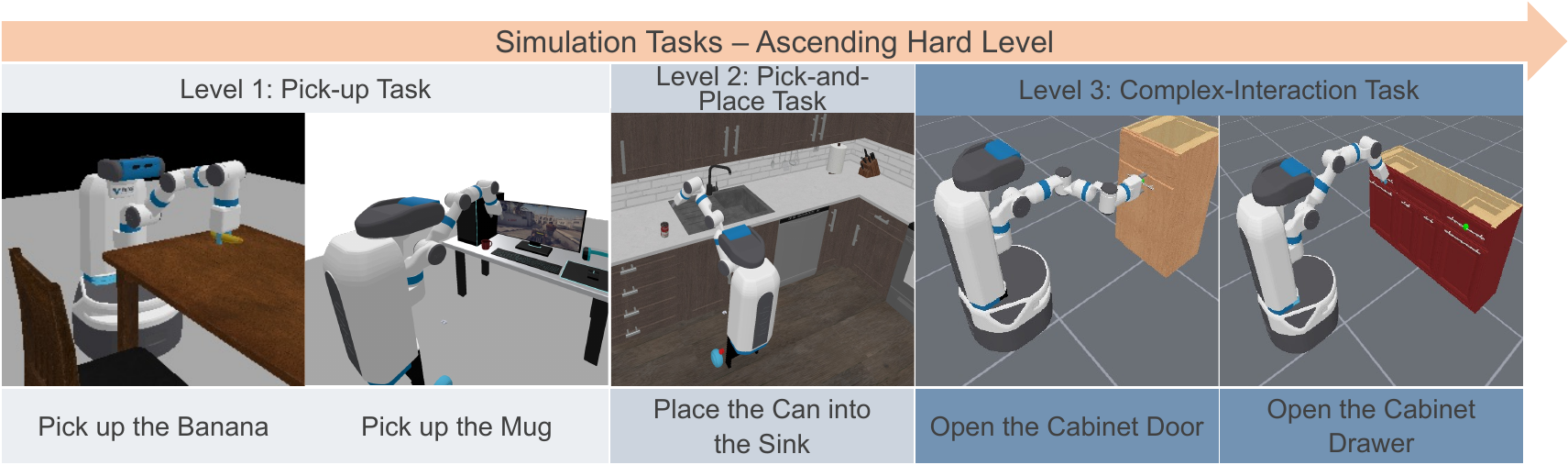} 
    \caption{Illustration of five simulated mobile manipulation tasks in ManiSkill environment with increasing difficulty: (1) \textbf{Pick-up tasks.} The robot navigates to a docking point and grasps the target object. (2) \textbf{Pick-and-place tasks.} The robot grasps the target object and place it into a designated container. (3) \textbf{Complex-interaction tasks.} The robot is required to interact with the handle to open the cabinet door or drawer without any collisions.}
    \label{fig:simtask}
    \vspace{-0.2cm}
\end{figure*}
\begin{table*}[t]
\centering
\small
\def\arraystretch{1.1}  %
\setlength\tabcolsep{7pt}  %
\caption{
ManiSkill success rates (SR, \%) and overall scores on one and five docking points. Overall SR is the average of all tasks}
\vspace{-0.3cm}
\scalebox{1}{%
\begin{tabular}{cccccccc}
\toprule
\multirow{2}{*}{\textbf{\#Dock}}& \multirow{2}{*}{\textbf{Method}} & \multicolumn{5}{c}{\textbf{Tasks Success Rates (\%)}$\uparrow$} & \multirow{2}{*}{\textbf{Overall SR (\%) $\uparrow$}} \\ \cline{3-7} 
 & & Pick Banana & Pick Mug & Place Can & Cabinet Door & Cabinet Drawer & \\ \midrule
\multirow{2}{*}{1} & DP & 95.0 & 82.0 & 76.0 & 68.0 & 72.0 & 78.6 \\
& DP3 & 100.0 & 88.0 & 90.0 & 81.0 & 84.0 & 88.6 \\
\midrule
\multirow{4}{*}{5} &DP & 19.0 & 16.4 & 15.6 & 13.6 & 14.4 & 15.8 \\
&DP3 & 20.0 & 17.7 & 18.8 & 16.2 & 16.4 & 17.8 \\
&DP3+DemoGen & \textbf{98.0} & 88.6 & 84.4 & 48.2 & 52.0 & 74.2 \\

& \cellcolor{gray!20}\textbf{DockAnywhere} & \cellcolor{gray!20}97.0 & \cellcolor{gray!20}\textbf{89.4} & \cellcolor{gray!20}\textbf{87.2} & \cellcolor{gray!20}\textbf{60.2} & \cellcolor{gray!20}\textbf{60.6} & \cellcolor{gray!20}\textbf{78.9} \\  

\bottomrule
\end{tabular}
}
\label{tab:sim}
\vspace{-0.3cm}
\end{table*}
For the instance in Fig.~\ref{fig:parse} (b), the \textit{skill} segment $\tau_1^{\text{s}}$ is preserved and the \textit{motion} segment $\tau_1^{\text{m}}$ is directly replanned given a closer docking point.

\subsection{Synthetic Observation Generation}
As shown in Fig.~\ref{fig:pipeline}, the observation consists of robot state and the representation of 3D point cloud, thus the synthetic observation generation can be divided into two parts: point cloud synthesis and robot state adaptation.

\noindent\textbf{Point Cloud Synthesis.} {Based on the segmentation results obtained in Sec.~\ref{sec:preprocess}, the geometric information of the robot and objects in the initial frame is identified. This enables precise relocation of the robot base by modifying the corresponding cluster of the point cloud }

{For the robot arm, we apply the spatial transformation according to the relative pose at \(p\) and the altered docking point \(p'\). Specifically, given the target SE(3) pose \(a_t^{\text{pose}}\) and \(\hat a_t^{\text{pose}}\) of the end-effector at \(p\) and \(p'\), respectively, we define the relative rigid transform:
\begin{equation}
    \Delta T_t = (a_t^{\text{pose}})^{-1}\hat a_t^{\text{pose}}
\end{equation}
Then the synthesized point cloud observation of the robot arm from \(p\) to \(p'\) is:
\begin{equation}
    \hat o_t^{\text{pc,arm}} = o_t^{\text{pc,arm}} \cdot \Delta T_t, \quad t \in [0,L)
\end{equation}
}

{For the manipulated objects, we assume that the object's point cloud $o_t^{\text{pc,obj}}$ remains unchanged during \textit{motion} segments. In contrast, during \textit{skill} segments, the object is visually merged with the robot arm. Therefore, to preserve the relative geometry between the robot arm and the manipulated object observed in the source trajectory, we apply the same rigid transformation to the object point cloud.}`

{The final synthesized point cloud observation is obtained by concatenating the transformed robot arm and object point clouds:
\begin{equation}
    \hat o_t^{\text{pc}} = [\hat o_t^{\text{pc,arm}}, \hat o_t^{\text{pc,obj}}], \quad t \in [0,L)
\end{equation}
where $[\cdot, \cdot]$ denotes the concatenation operation.}

\noindent\textbf{Robot State Adaptation.} The robot state consists of the end-effector state $o_t^{\text{pose}}$ and the gripper state $o_t^{\text{cmd}}$. Similar to the action generation, the gripper and end-effector state is invariant and equivariant to the same spatial transformation, respectively. Therefore, the synthesized robot state $\hat o_t^{\text{pose}}$ and $\hat o_t^{\text{cmd}}$ can be formulated as:
\begin{align}
\begin{split}
    \hat o_t^{\text{cmd}} &= o_t^{\text{cmd}}, \quad t \in [0, L) \\
    \hat o_t^{\text{pose}} & = o_t^{\text{pose}} \cdot (a_t^{\text{pose}})^{-1} \cdot \hat a_t^{\text{pose}}, \quad t \in [0, L)
\end{split}
\end{align}

Finally, we obtain the synthesized visual and state observation $\hat o_t = (\hat o_t^{\text{pc}}, \hat o_t^{\text{pose}}, \hat o_t^{\text{cmd}})$ along with the generated action $\hat a_t = (\hat a_t^{\text{pose}}, \hat a_t^{\text{cmd}})$ to form the augmented demonstration $\mathcal D_{p_i}$ for the novel docking point $p_i$.

\begin{figure*}
\centering
\includegraphics[width=0.8\textwidth]{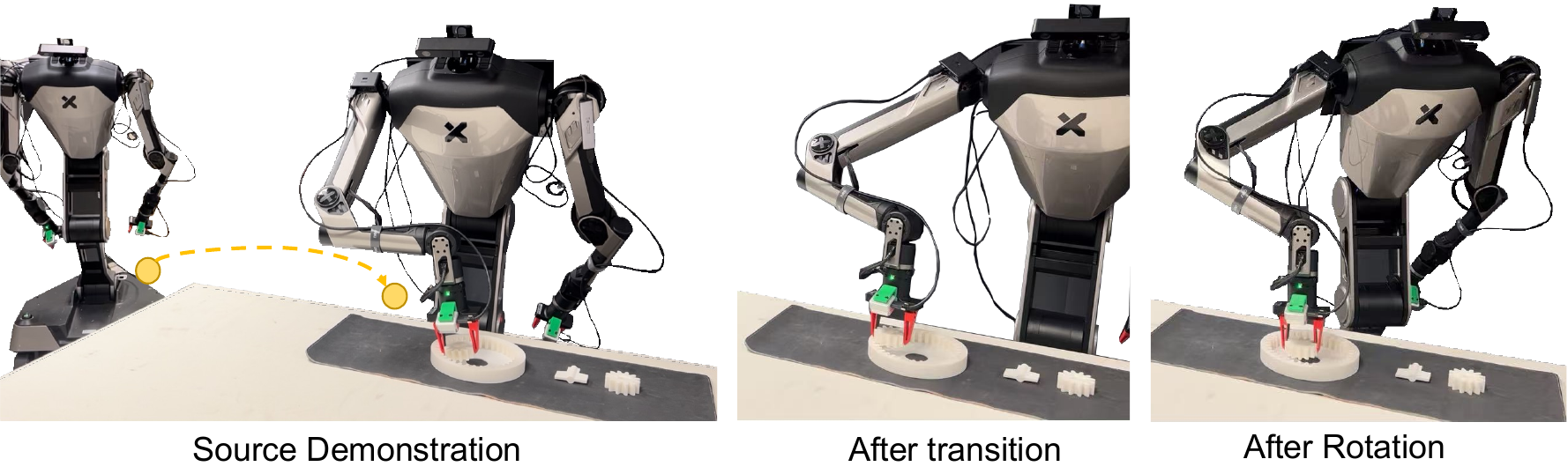}
\vspace{-0.3cm}
\caption{Real-world experiment visualization for task of placing gear onto plate. Source demonstration (left) in real world experiment of navigating and placing gear. Trained policy still works after transition (middle) and rotation (right) to simulate suboptimal docking points.}
\end{figure*}

\begin{table*}[t]
\centering
\small
\caption{Ablation studies on the number of additional docking points for augmentation and training the policy. The results are tested on an additional unseen docking point.}
\setlength{\tabcolsep}{11pt}
\begin{tabular}{ccccccc}
\toprule
\textbf{\#Dock} & \textbf{Pick Banana} & \textbf{Pick Mug} & \textbf{Place Can} & \textbf{Cabinet Door} & \textbf{Cabinet Drawer} & \textbf{Overall SR (\%) $\uparrow$} \\ \midrule
 1 & 45.0 & 38.2 & 35.6 & 30.2 & 28.4 & 35.5 \\
 2 & 58.4 & 52.6 & 48.2 & 42.4 & 40.8 & 48.5 \\       
 \cellcolor{gray!20}4 (ours) & \cellcolor{gray!20}72.6 & \cellcolor{gray!20}68.4 & \cellcolor{gray!20}65.2 & \cellcolor{gray!20}52.8 & \cellcolor{gray!20}50.6 & \cellcolor{gray!20}61.9 \\  
 6 & 75.2 & 70.8 & 67.4 & 54.2 & 52.0 & 63.9 \\  
 \bottomrule
\end{tabular}
\vspace{-0.45cm}
\label{tab:ablation}
\end{table*}

\section{EXPERIMENTS}

In this section, we evaluate the effectiveness of \cmrword{DockAnywhere} framework through a series of experiments designed in ManiSkill \cite{taomaniskill3}. 
We first formulate the details of mobile manipulation tasks created to simulate realistic scenarios subject to non-ideal or suboptimal docking points, on which we compare the performance of \cmrword{DockAnywhere} on these tasks against static manipulation baselines and augmentation methods.
Subsequently, we validate the individual components of \cmrword{DockAnywhere} through ablation studies.
Finally, we demonstrate the generalization capabilities by deploying the trained policies on real-world environments.

\subsection{Task Setup and Implementation Details}
\noindent\textbf{Task Setup.} We design five mobile manipulation tasks in ManiSkill \cite{taomaniskill3}, a simulation benchmark built upon SAPIEN, offering diverse scene datasets (\textit{e.g.,} RoboCasa \cite{nasiriany2024robocasa}) and official manipulation tasks. as illustrated in Fig. \ref{fig:simtask}. The tasks are defined in increasing difficulty to comprehensively evaluate the performance of mobile manipulation policies.

\noindent\textbf{Implementation Details.} We use DP3 \cite{Ze2024DP3} as our policy head. To train the head, we use the keyboard teleoperation to collect 10 source demonstrations at a feasible docking point for each task. The policy is trained for 2000 epochs on a single NVIDIA A6000 GPU.
During training, we generate augmented demonstrations for five assigned docking points to simulate perturbated docking points or  non-ideal points computed by a navigation module. The five docking are randomly sampled within a reasonable range, considering both the distance and orientation from the target object. {The distance threshold between the end-effector and target object is set to 0.1m, since it is a regular manipulation range for table-top objects.}

\subsection{Comparison with Baselines}
We compare \cmrword{DockAnywhere} with the vanilla DP3 and 2D diffusion policy (DP) as static manipulation baselines. Additionally, we implement DemoGen \cite{xue2025demogen} as a data augmentation baseline, which synthesizes visual observations using affine transformations. The methods are evaluated on both one and five docking points, where the five docking points are within a reasonable range, considering both the distance and orientation from the target object. For the single docking point setting, the point is ensured to be feasible. The results are in Table \ref{tab:sim}.

As shown in Table \ref{tab:sim}, though DP and DP3 may achieve satisfactory performance when tested on the single docking point, they suffer from severe performance degradation when tested on five docking points, with the overall success rate dropping to 15.8\% and 17.8\%, respectively. This degradation further demonstrates the view generalization problem, where the policies fail to adapt to the diverse viewpoints caused by unseen docking points.
In addition, DemoGen performs worse on the complex-interaction tasks where the robot's end-effector needs significant pose adaptation (\textit{e.g.,} pose changing for opening the cabinet drawer when orientation varies) for novel docking points. 
{DemoGen only achieves a success rate of 40\% on a rotated point for simple pick-mug task, while our method presents 100\% success rate, which further validates the importance of our novel augmentation space with full SE(3) spatial variations.}
{Finally, our \cmrword{DockAnywhere} achieves the best performance across all tasks and docking points, by explicitly modeling docking-point variation together with the relative robot–object geometry, including both translation and rotation, whereas DemoGen mainly considers approximately parallel object-pose changes on tabletop scenarios.}
\begin{table}[t]
\centering
\small
\caption{{Success rates (\%) under different distances between the robot base and the target object ranging from 0.4\,m--0.8\,m.}}
\setlength{\tabcolsep}{27pt}
\begin{tabular}{cc}
\toprule
\textbf{Range (ratio)} & \textbf{Success Rate (\%)} \\ \midrule
0.5 & 65.6 \\
0.8 & 86.4 \\
\cellcolor{gray!20}1.0 & \cellcolor{gray!20}87.2 \\
2.0 & 46.3 \\ 
\bottomrule
\end{tabular}
\label{tab:range}
\vspace{-0.4cm}
\end{table}
\subsection{Ablation Studies}
In this section, we verify the effectiveness of our \cmrword{DockAnywhere} framework through ablation studies in different docking point configurations. Specifically, we investigate the impact of the number of additional docking points for augmentation and the spatial range of these points.

To study the influence of the number of additional docking points, we vary the number of docking points from 1 to 6 and evaluate the trained policy on an unseen docking point during training. The results are shown in Table \ref{tab:ablation}. It can be observed that as the number of docking points increases, the success rate consistently improves across all tasks. This indicates that augmenting more docking points during training enhances the policy's ability to generalize to unseen docking points. However, the performance gain diminishes (\textit{i.e.,} 2\% gain from 4 to 6 additional docking points) as the number of docking points increases, we set the number of augmented docking points as four to consider the trade-off between policy performance and the computational cost.

{The spatial distance of the augmented docking point is a critical factor for the final performance. To investigate this, we vary the spatial range of the docking points from 0.5 to 2.0 times the source docking distance and evaluate the trained policy on an unseen docking point during training for the place-can task. The range ratio is defined as the distance between the robot base and the target object at the augmented docking point divided by that at the source docking point. In our tasks, the source distance is approximately 0.4\,m--0.8\,m depending on the task setup}. The results are shown in Table \ref{tab:range}. It can be observed that as the range increases, the success rate first increases and then decreases, indicating that a moderate range is beneficial for generalization, while too large a range may lead to excessive perturbation and hinder performance.

\subsection{Real World Experiments}
\begin{figure}[t]
    \centering
    \includegraphics[width=0.85\linewidth]{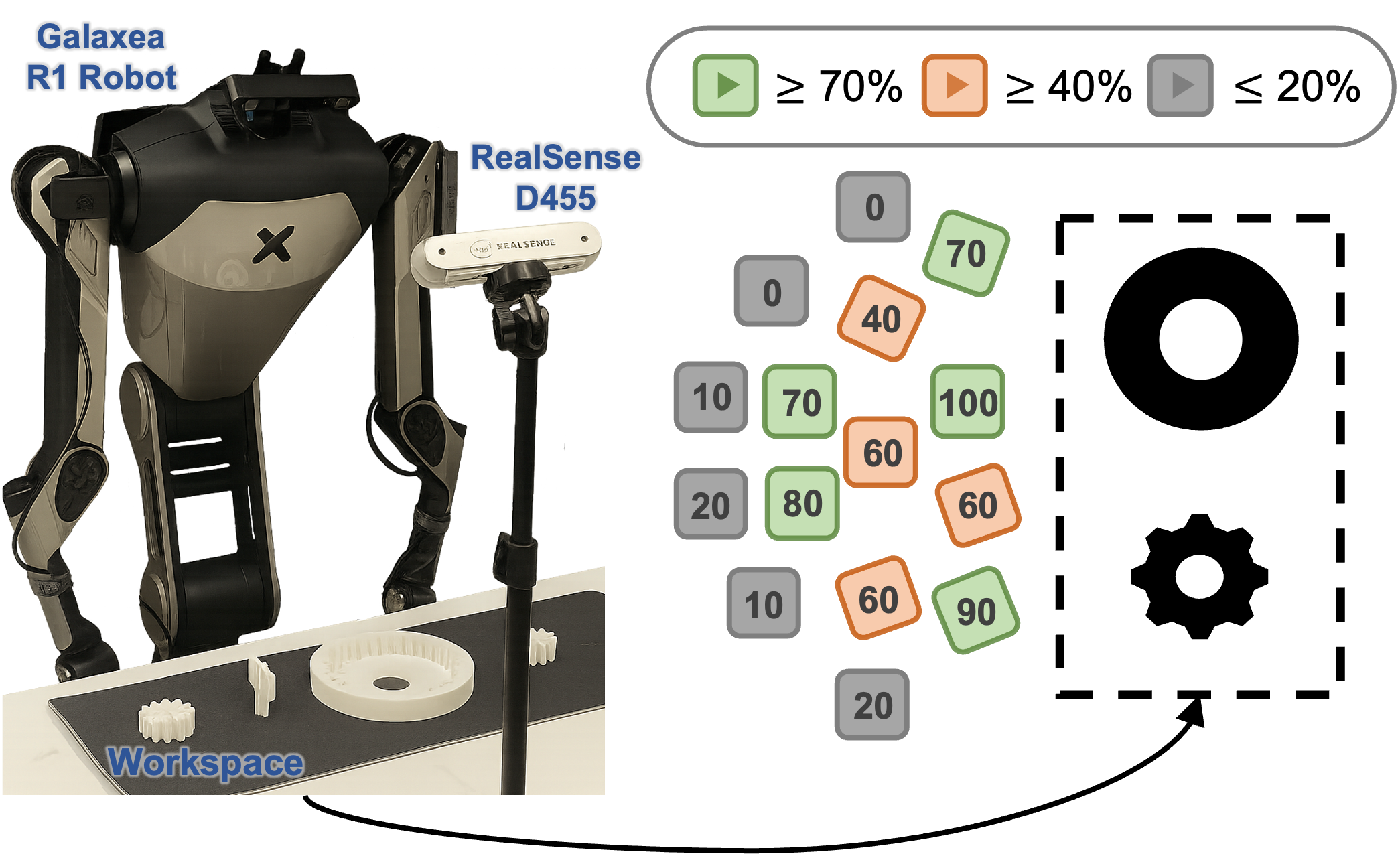}
    \caption{Real-world task setup (left) and success rates at different docking points with varying ranges for task of placing gear onto plate (right). Rotated squares indicate the orientation of the robot. Augmented docking points for training is in green color.}
    \label{fig:real_ablation}
    \vspace{-0.3cm}
\end{figure}

We deploy the trained policies on a Galaxea R1 mobile manipulator consisting of two A1 6-DoF arms and an omnidirectional mobile base, a ZED2 depth camera and a Livox LiDAR sensor for
accurate localization. Objects within the scene are segmented into individual instances using GroundedSAM \cite{ren2024grounded}.
A realsense D455 RGBD camera is used for the only visual observation input to the policy. The setup of real-world experiments is illustrated in Fig. \ref{fig:real_ablation}.
\ans{The performance results are summarized in Table  \ref{tab:real_main}.} We evaluate the policies on three representative popular manufacturing tasks: picking up gear, placing gear onto plate, and assembling components. The performance is evaluated at an unseen docking point after training with one source demonstration and four augmented docking points. The trained policies achieve success rates of 60\%, 40\%, and 30\% on these three tasks, respectively. 
\begin{table}[t]
\centering
\small
\caption{Success rates (\%) of three real-world tasks.}
\setlength{\tabcolsep}{19pt}
\begin{tabular}{cc}
\toprule
\textbf{Real-World Task} & \textbf{Success Rate (\%)} \\ \midrule
Picking up gear & 60 \\
Placing gear onto plate & 40 \\
Assembling components & 30 \\
\bottomrule
\end{tabular}
\label{tab:real_main}
\vspace{-0.6cm}
\end{table}

{Furthermore, to comprehensively investigate the generalization capability of the trained policy across different docking points for testing, we conduct ablation studies by varying the location of the unseen docking point, as shown in Fig.~\ref{fig:real_ablation}. For each docking-point condition, we conduct 10 independent trials and compute the success rate accordingly.}
The results indicate that: (1) The trained policy consistently performs well at the augmented docking points used for training, demonstrating the effectiveness of \cmrword{DockAnywhere} in improving policy performance at these locations. (2) the success rate generally decreases as the docking point moves further away from the range of augmented points , which aligns with our intuition that larger spatial perturbations pose greater challenges for the policy. Notably, even at a significant out-of-distribution distance (\textit{e.g.,} 0.4m) from the target object, the policy still achieves a success rate around 20\%, demonstrating its robustness to navigation errors and non-ideal reachability in real-world scenarios. 
This real-world ablation study further reveals the potential of \cmrword{DockAnywhere} to address the view generalization problem caused by docking point perturbation in a synthetic and low-cost manner, provided by sufficient augmented points covering a certain spatial range for training.

\begin{figure}
    \centering
    \includegraphics[width=1\linewidth]{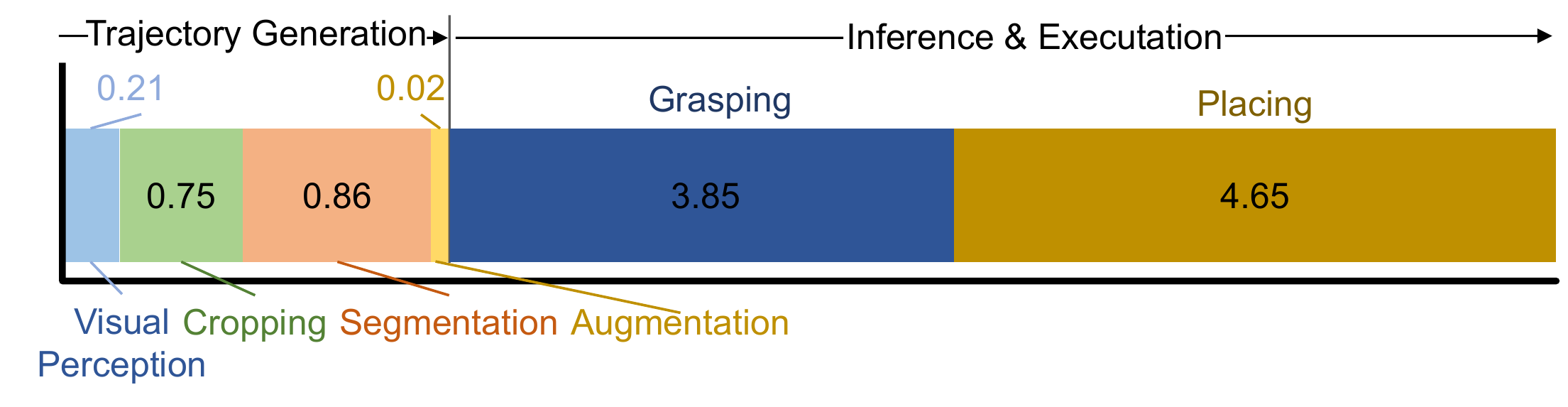}
    \vspace{-0.6cm}
    \caption{Time used for augmenting trajectory, inference and execution. Measured on real-world task of placing gear onto plate.}
    \label{fig:time}
    \vspace{-0.5cm}
\end{figure}

Finally, as shown in Fig.~\ref{fig:time}, we compare the used time for generating a single novel trajectory in the training stage, inference and executing actions in task of placing gear onto plate. Fig.~\ref{fig:time} reveals the minimal time cost (\textit{i.e.}, 0.02 seconds for a single augmentation) of generating new demonstrations of our \cmrword{DockAnywhere} compared to the inference and action execution stage, showing the significant time efficiency of our method.

\section{CONCLUSION}
{In this paper, we presented \cmrword{DockAnywhere}, a data-efficient framework to improve the viewpoint generalization capability of mobile manipulation. We address viewpoint generalization in mobile manipulation by lifting a single demonstration to diverse feasible docking configurations. By decoupling docking-dependent base motions from invariant skills, the proposed approach expands training coverage without additional real-world data. The generated trajectories and observations remain spatially consistent through 3D synthesis, enabling policies trained on minimal data to generalize reliably to unseen docking points in both simulation and real-world settings.
Extensive experiments on ManiSkill and real-world platform demonstrate that \cmrword{DockAnywhere} substantially improves policy performance and reveal that trained policies can successfully adapt to navigation errors and non-ideal docking locations. Overall, \cmrword{DockAnywhere} offers a practical and efficient solution for enhancing the view generalization capability of mobile manipulation systems.}
\bibliographystyle{IEEEtran}
\bibliography{ref}

\addtolength{\textheight}{-12cm}

\end{document}